\documentclass{pprai}

\pdfoutput=1
\usepackage[utf8]{inputenc}

\usepackage[T1]{fontenc}
\usepackage{graphicx}
\usepackage{amsthm}
\usepackage{txfonts}
\usepackage{hyperref}
\usepackage{url}
\usepackage{float}
\usepackage{caption}
\usepackage{subcaption}
\usepackage{algpseudocode}
\usepackage[dvipsnames]{xcolor}
\usepackage{pdfpages}

\title{HeROS: a~miniaturised platform for research and development on Heterogeneous RObotic Systems}
\headtitle{HeROS: a~miniaturised platform for Heterogeneous RObotic Systems}

\author{Tomasz Winiarski$^{[0000-0002-9316-3284]}$, \\ Daniel Giełdowski$^{[0000-0002-4348-2981]}$,\\ Jan Kaniuka$^{[0009-0009-9379-4906]}$,\\ Jakub Ostrysz$^{[0009-0006-8178-0134]}$,\\ Jakub Sadowski}
\headauthor{T. Winiarski, D. Giełdowski, J. Kaniuka, J. Ostrysz, J. Sadowski}
\affiliation{%
	 Warsaw University of Technology\\
	 Faculty of Electronics and Information Technology\\
	 Institute of Control and Computation Engineering\\
	 Nowowiejska 15/19, 00-665 Warsaw, Poland\\
	 tomasz.winiarski@pw.edu.pl}

\keywords{HeROS, robotics, miniaturised physical test-bed}

\begin{document}
	
	\includepdf{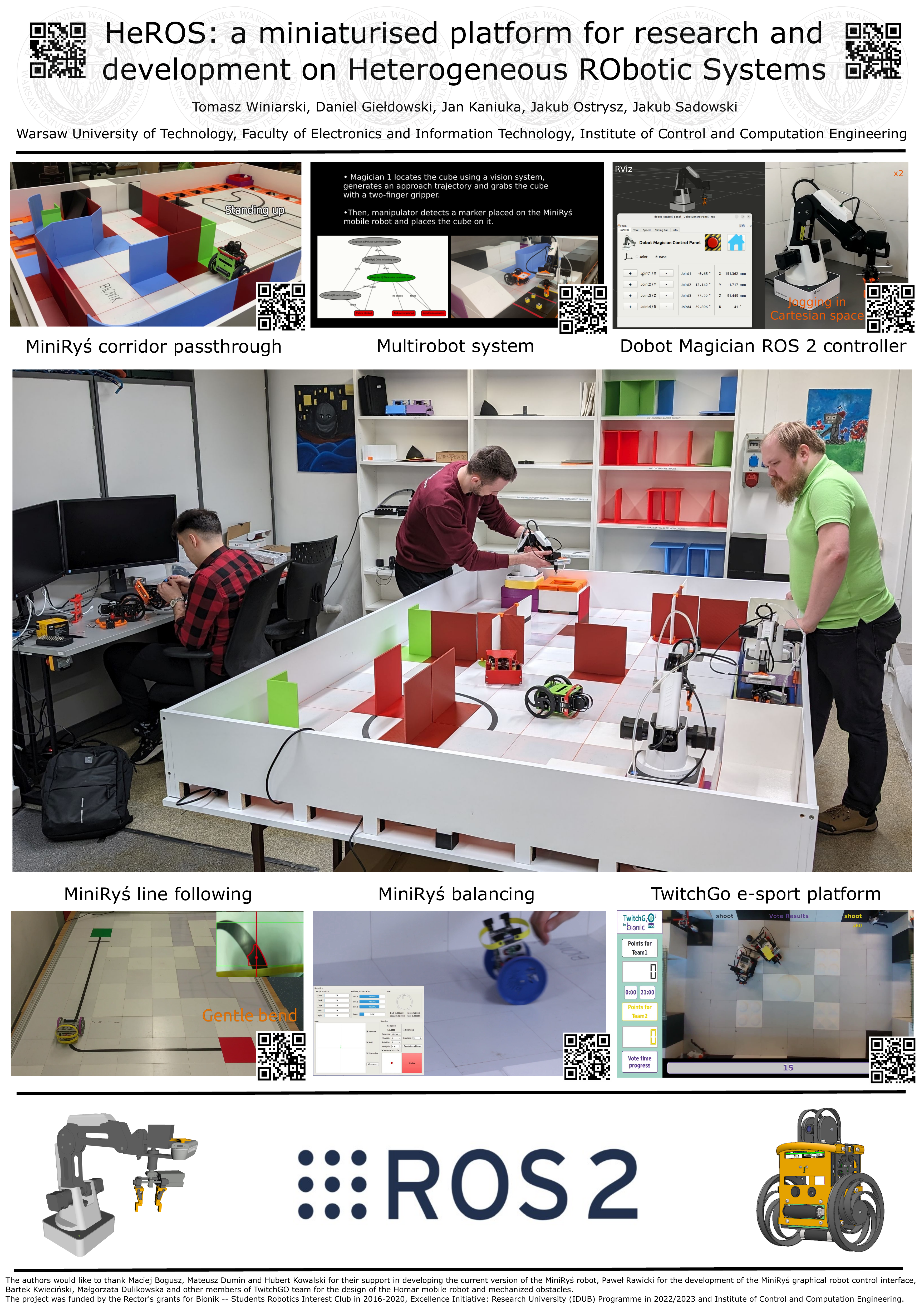}
	
	\maketitle
	
	\begin{abstract}
		Tests and prototyping are vital in the research and development of robotic systems. Work with target hardware is problematic. Hence, in the article, a~low-cost, miniaturised physical platform is presented to deal with experiments on heterogeneous robotic systems. The platform comprises a~physical board with tiles of the standardised base, diverse mobile robots, and manipulation robots. The number of exemplary applications validates the usefulness of the solution.
	\end{abstract}
	
	\section{Introduction}
	\label{sec:introduction}
			
	Effective development and testing of robotic systems is a~vast topic. Physical test platforms are an essential aspect thereof \cite{dudek2023spsysml}. Running tests on target hardware is problematic, if only because of its cost and availability and the risks associated with the ongoing work. 
	In the case of modern robotic systems, this problem is particularly relevant, as artificial intelligence algorithms are currently used and intensively investigated in robotics \cite{shaukat2020impact}, that introduces many general and specific risk factors \cite{cheatham2019confronting}.		
			
	This paper proposes a~miniaturised, low-cost physical test platform for heterogeneous robotic systems. The nature of the platform and its requirements are presented in sec.~\ref{sec:requirements}. The platform's components developed so far are described in sec.~\ref{sec:components}, and example applications are described in sec.~\ref{sec:applications}. Section~\ref{sec:conclusions} concludes the work.
		
	\section{Requirements}
	\label{sec:requirements}
	
	\begin{itemize}
		\itemsep -0.3em 
		\item [R1 -] modularised, tile-based board that makes it easy to reproduce experiments in various configurations of the environment,
		\item [R2 -] small enough tiles to build complicated test environments in regular building premises,
		\item [R3 -] easy access to the inside of the board from all sides,
		\item [R4 -] moderate board price with low-scale manufacturing,
		\item [R5 -] easy to produce and modify,
		\item [R6 -] the possibility of routing cables underneath the tiles,
		\item [R7 -] mobile robots that fit the dimensions of the board tiles,
		\item [R8 -] manipulation robots that fit the dimensions of the board tiles,
		\item [R9 -] common software framework for all of the hardware.
	\end{itemize}

	\section{Components of the platform}
	\label{sec:components}
	
	\subsection{Board and tiles}
    \label{subsec:board_and_tiles}

\begin{figure}[h]
        \begin{subfigure}[b]{0.32\linewidth}
            \centering
            \includegraphics[width=\linewidth]{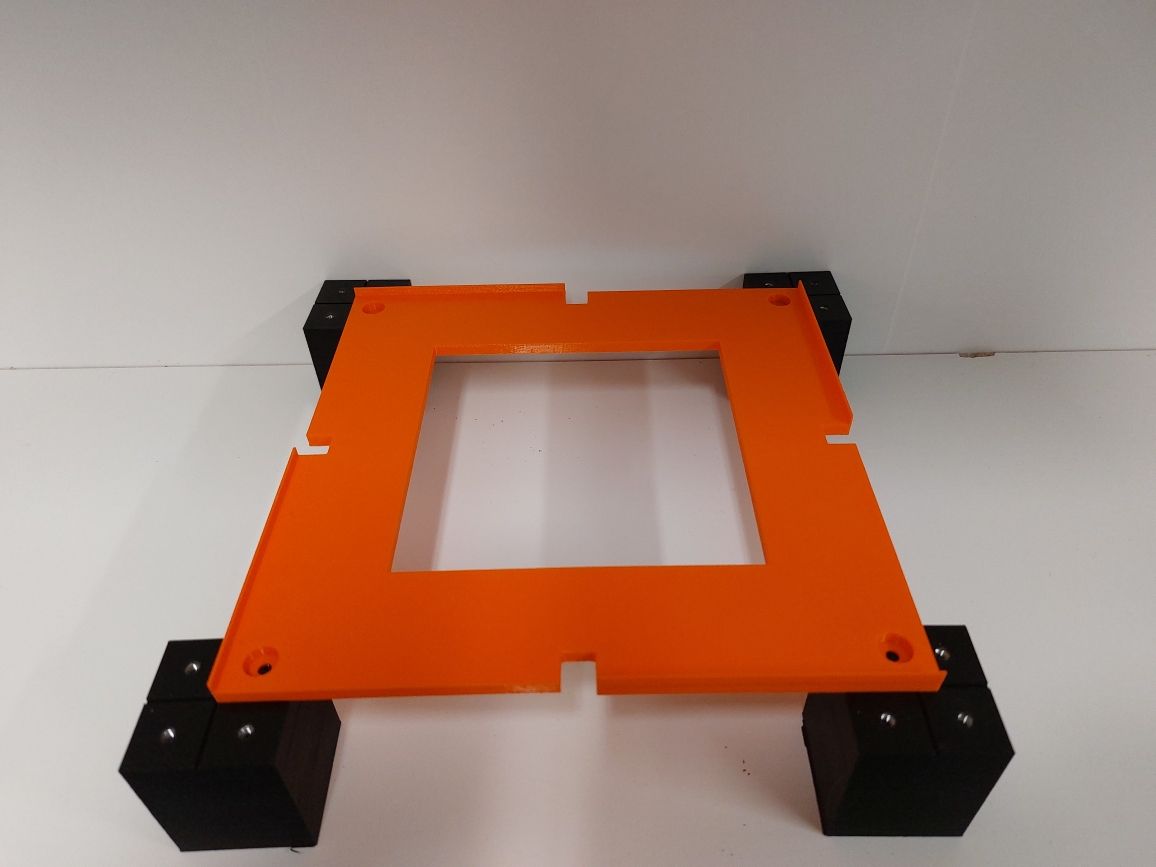} 
            \caption{Single board base standing on legs} 
            \label{subfig:board_base} 
        \end{subfigure}
        \begin{subfigure}[b]{0.32\linewidth}
            \centering
            \includegraphics[width=\linewidth]{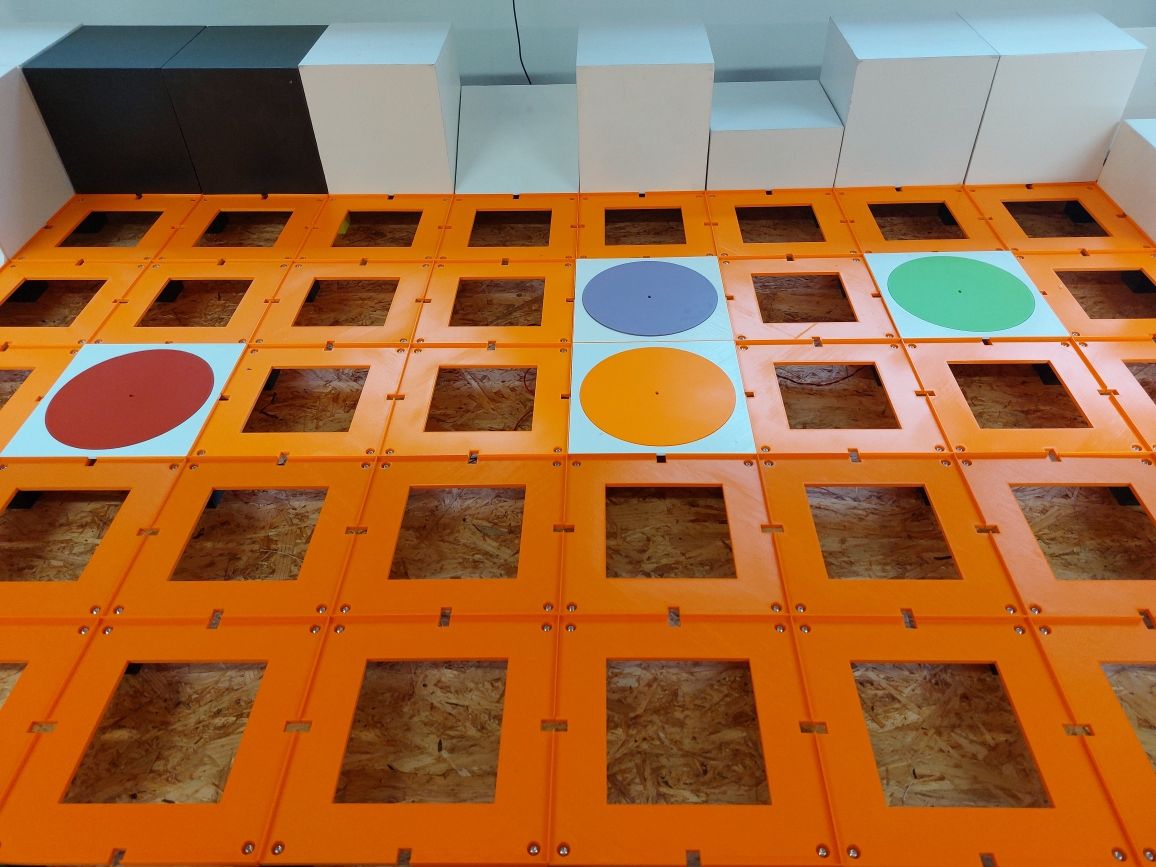} 
            \caption{Smaller board, partly filled} 
            \label{subfig:board_bare} 
        \end{subfigure}
        \begin{subfigure}[b]{0.32\linewidth}
            \centering
            \includegraphics[width=\linewidth]{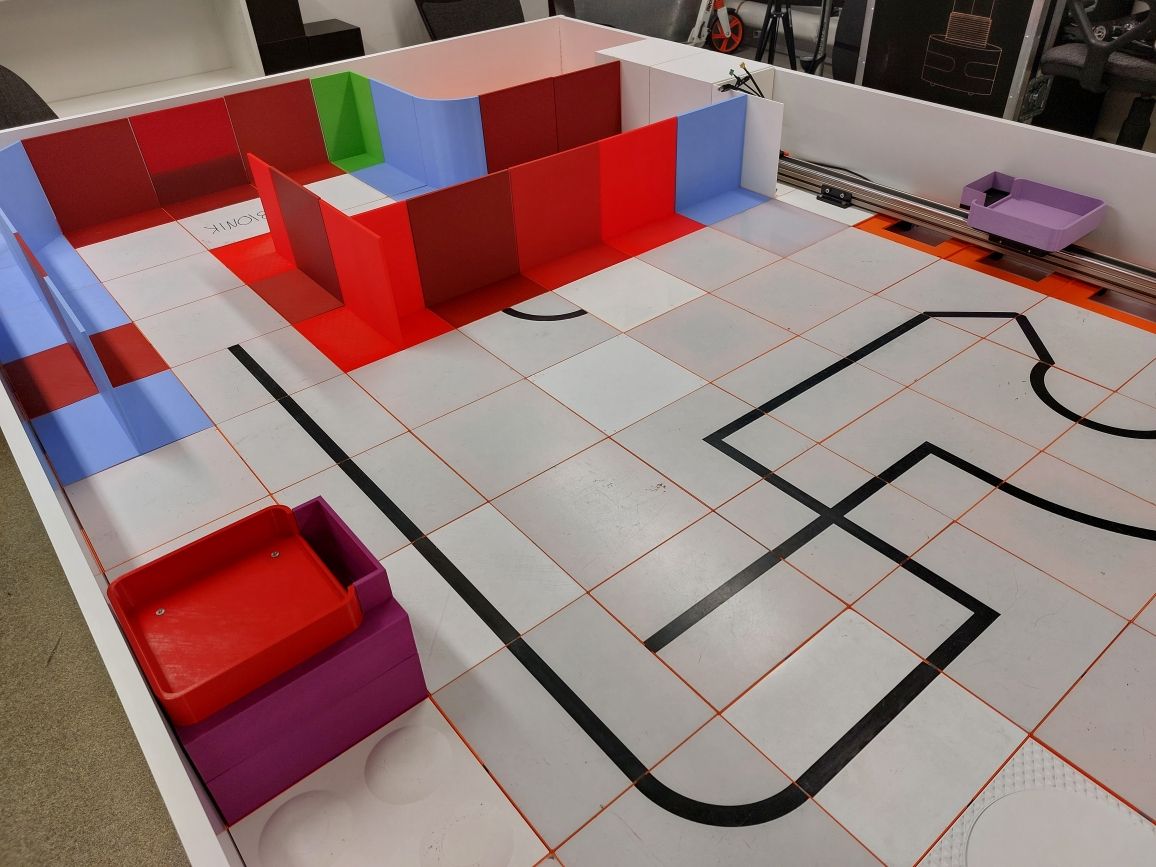} 
            \caption{Larger board with different tiles present} 
            \label{subfig:large_board} 
        \end{subfigure}
        \caption{HeROS board}
        \label{fig:heros_board} 
    \end{figure}

   In order to satisfy [R1], the platform utilizes a~modular tile-based board. The board is a~successor to the environment described in \cite{par-plansza-eng}. It is constructed from the following parts: legs, bases and tiles. The board's legs are 5cm high to allow the placement of vulnerable electrical components (cables, chargers, motors, compressors, etc.) below the robotic environment as per [R6]. Each base is placed on four legs and can fit a~single tile (fig. \ref{subfig:board_base}). The tiles have square-shaped bases, 20cm in side size. Currently, we designed several different tiles, including flat floor tiles, tiles with walls, turns and bends. We recently included proper bases and a~sliding rail for the manipulation robot. This allows for building complex testing environments in finite space while consistent with previous tile designs [R2]. Until now, two versions of the environment were prepared, fitting six by ten (fig.~\ref{subfig:board_bare}) and eight by twelve tiles (fig. \ref{subfig:large_board}). Both allow easy access to the whole board, satisfying requirement [R3]. The whole board was created using 3D printing in order to make it both easy [R5] and relatively cheap [R4] to produce and modify.
    
	\subsection{Mechanised obstacles}
    \label{subsec:mechanised_obstacles}

    To improve the variety of experiments, we designed some mechanised obstacles. The obstacles extend the board with hazards [R1]. Currently, we have one functional obstacle -- spinning tile (fig. \ref{subfig:spinning_wheel}) -- and one in the phase of model improvements -- moving wall (fig. \ref{subfig:moving_wall}). A~linear motor propels the spinning wheel and induces rotation to the robot placed upon it. The moving wall uses the stepper motor to change the board's structure while robots navigate it.
	
	\subsection{Homar mobile robot}
    \label{subsec:homar_mobile_robot}

\begin{figure}[h]
        \begin{subfigure}[b]{0.32\linewidth}
            \centering
            \includegraphics[width=\linewidth]{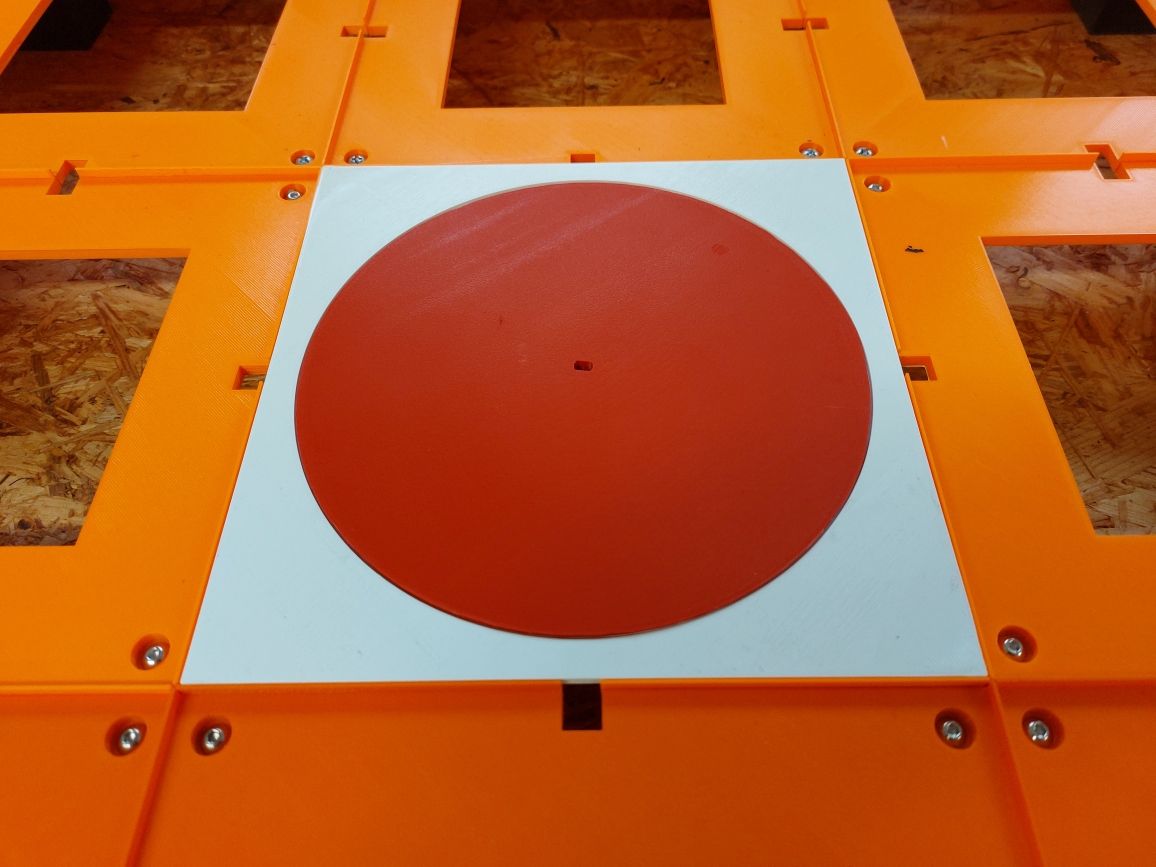} 
            \caption{Spinning wheel obstacle} 
            \label{subfig:spinning_wheel} 
        \end{subfigure}
        \begin{subfigure}[b]{0.32\linewidth}
            \centering
            \includegraphics[width=\linewidth]{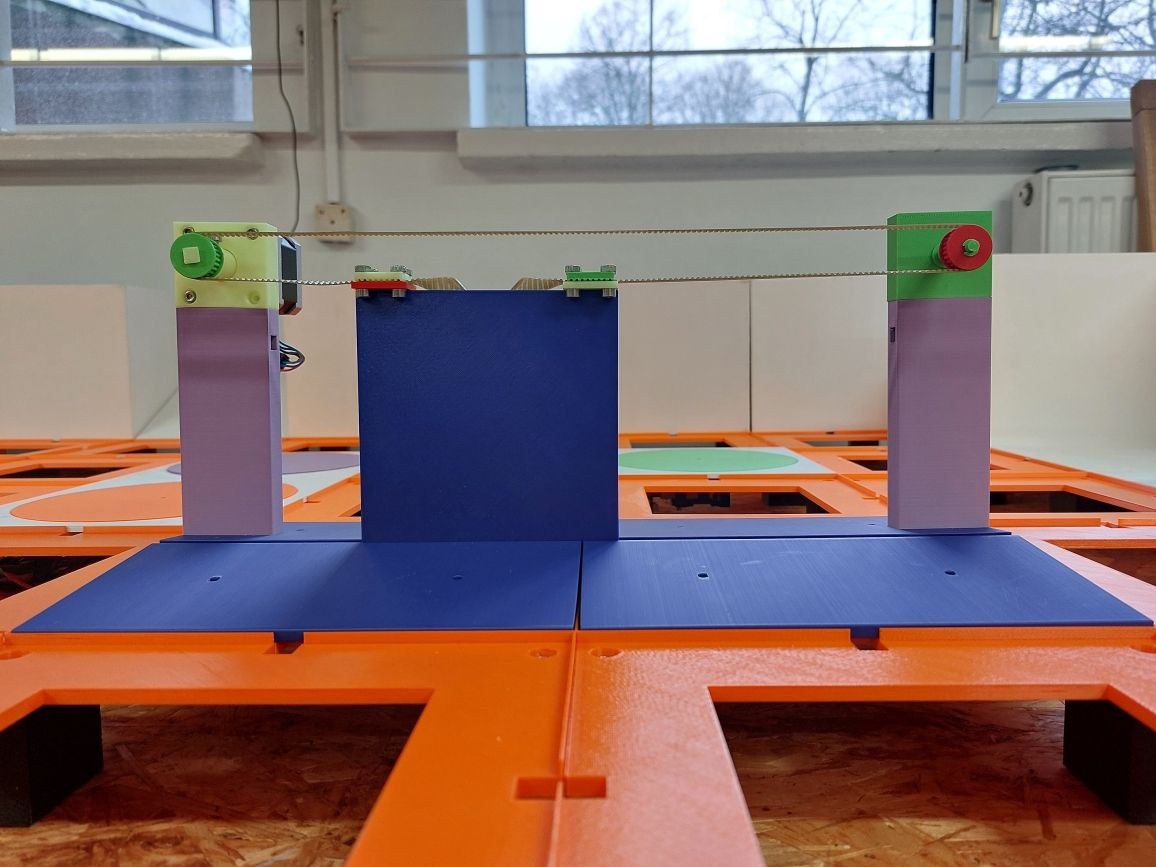} 
            \caption{Moving wall obstacle} 
            \label{subfig:moving_wall} 
        \end{subfigure}
        \begin{subfigure}[b]{0.32\linewidth}
            \centering
            \includegraphics[width=\linewidth]{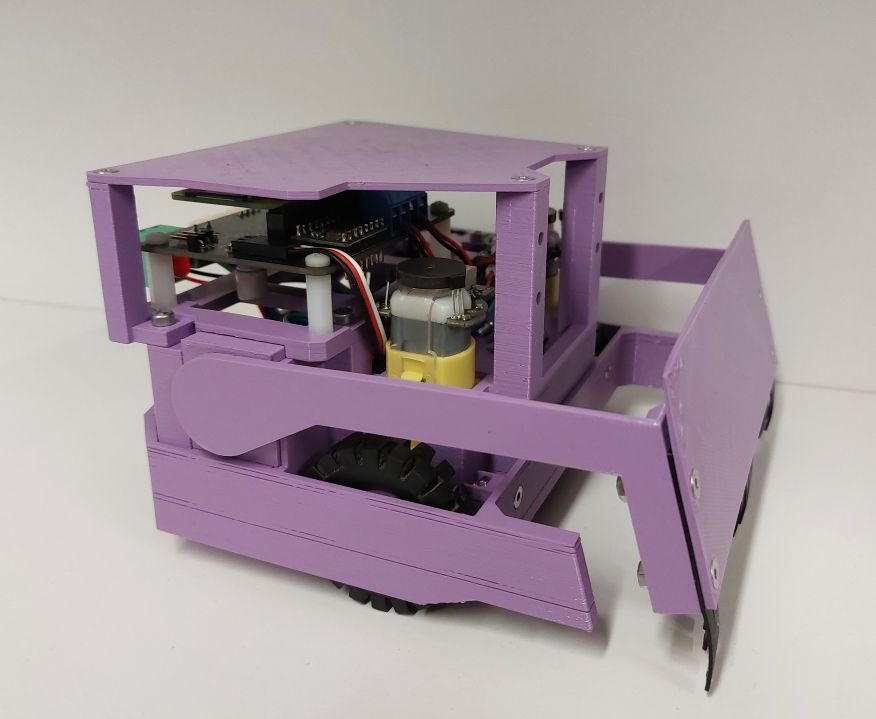} 
            \caption{Homar mobile robot} 
            \label{subfig:homar} 
        \end{subfigure}
        \caption{HeROS board}
        \label{fig:obstacles and homar} 
    \end{figure}

 Homar robot (fig. \ref{subfig:homar}) is a~platform developed in a~student-driven project briefly described in sec. \ref{subsec:twitchgo}. Its blueprint fits roughly in a~15 cm-sized square, making it possible to operate within a~single tile [R7]. It is a~mobile robot with differential-based and front-wheel drive with limited capabilities of affecting the environment using a~built-in shovel. This is achieved using two linear motors for movement and a~single servomotor for its tool. The robot is also equipped with an onboard Raspberry Pi Zero computer, a~custom motherboard, and two encoder sensors. Its control system is integrated with ROS 2 libraries for software unification [R9].

\subsection{Customised Dobot Magician robot}

  Dobot Magician is a~4-DOF serial educational manipulator manufactured by the Dobot company. Different tools can be attached to the robot, such as a~two-finger gripper, a~pneumatic suction cup and a~soft gripper. As part of the work of the Bionik - Students Robotics Interest Club, the robot was equipped with additional 3D-printed accessories such as gripper extenders, a~mount for the Intel RealSense D435i depth camera and mounts for signal cables and a~cable from the compressor. A~ROS 2-based control system with a~graphical interface has been developed for Dobot Magician~\cite{jkaniuka-bsc-23-eng}. The kinematic structure of this robot makes it suitable for typical \textit{Pick}\&\textit{Place} tasks.

     \begin{figure}[!htb]
    	\centering
    	\def\myheight{5cm}
    	\subfloat[\label{fig:kształtowy}]{\includegraphics[height=\myheight]{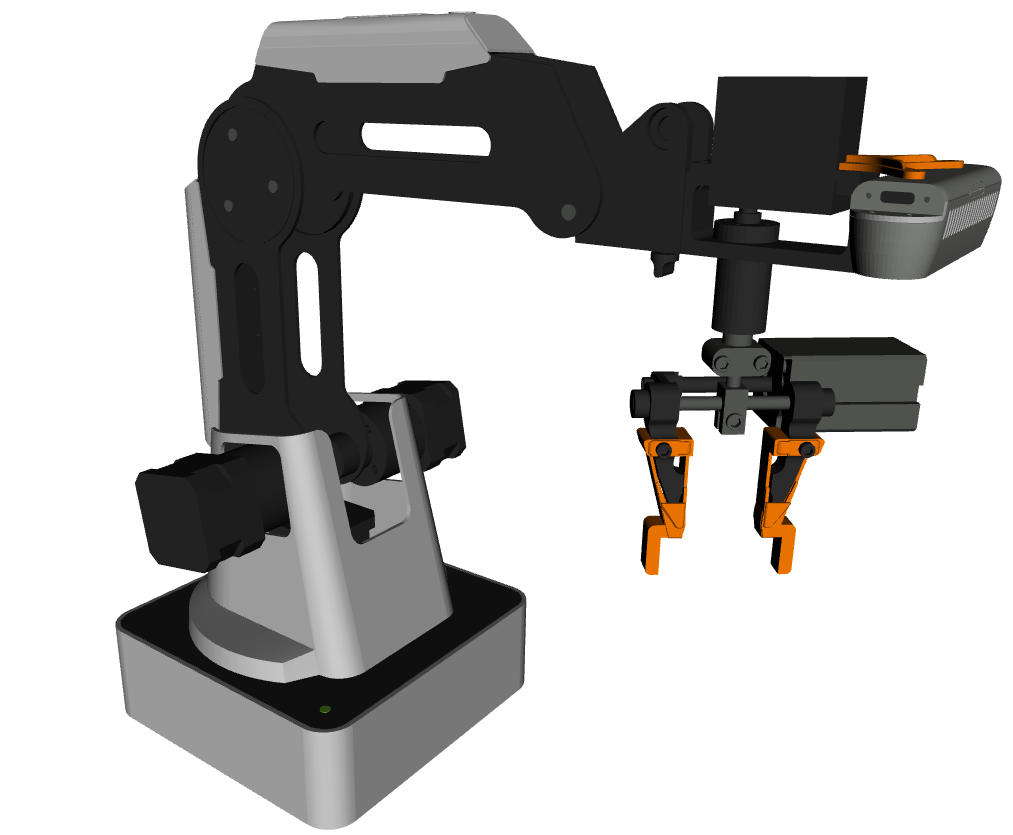}\label{fig:dobot}} \quad \quad \quad \quad 
    	\subfloat[\label{fig:collision}]{\includegraphics[height=\myheight]{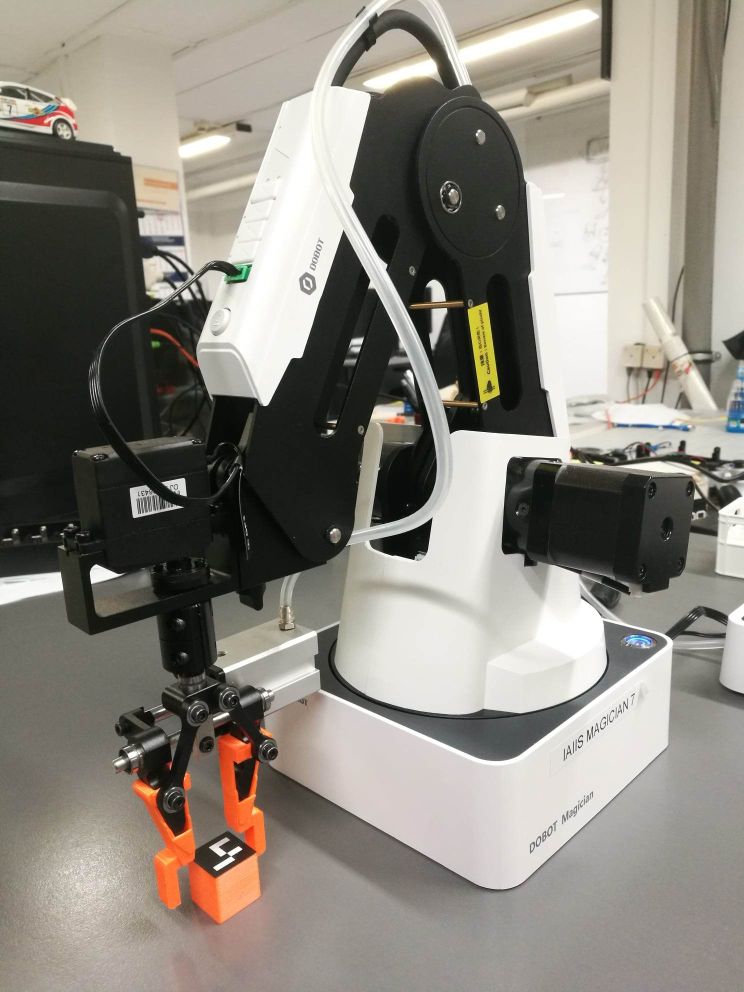}\label{fig:rys}} \quad
    	\caption{Customised Dobot Magician robot - visualization and real view.}
    \end{figure}
 
	\subsection{MiniRyś mobile robot}
 
\begin{figure}[h]
        \begin{subfigure}[b]{0.5\linewidth}
            \centering
            \includegraphics[width=0.6\linewidth]{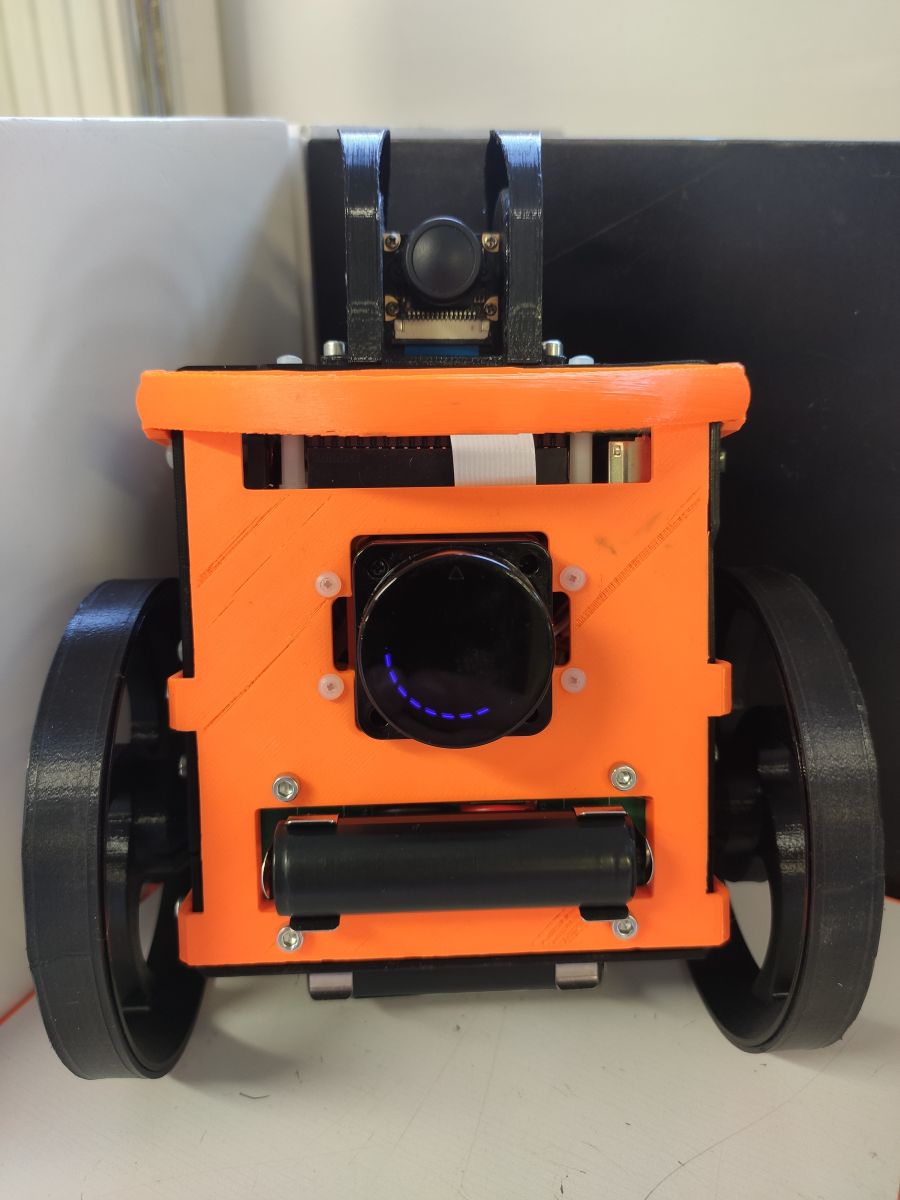} 
            \caption{with LiDAR} 
            \label{fig:MRlidar} 
            \vspace{4ex}
        \end{subfigure} 
        \begin{subfigure}[b]{0.5\linewidth}
            \centering
            \includegraphics[width=0.6\linewidth]{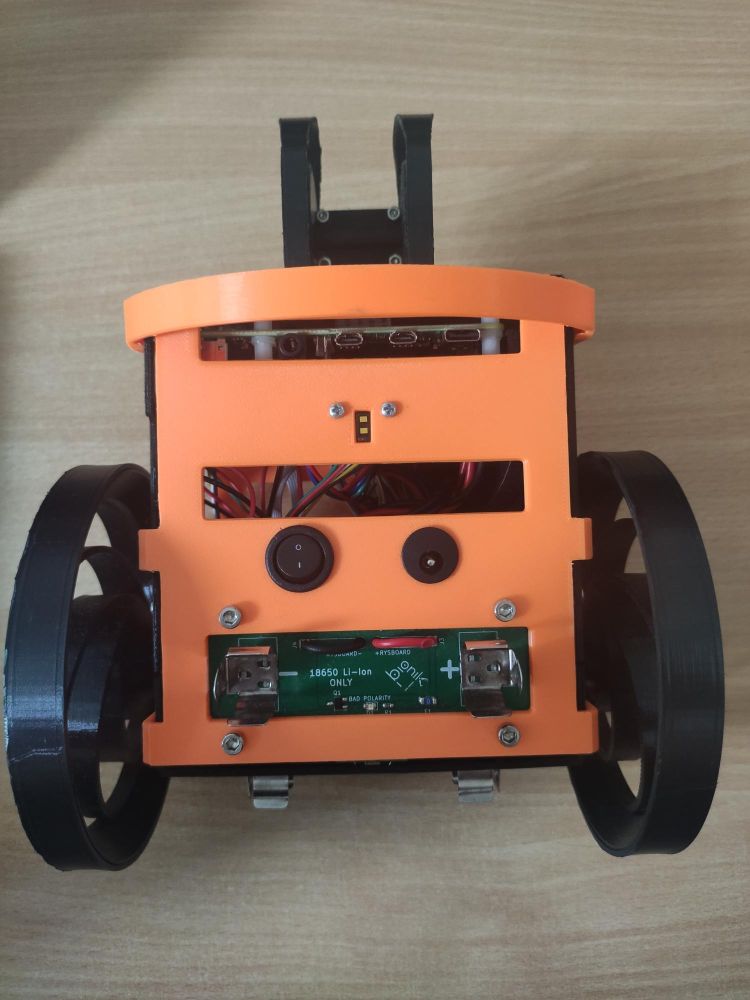} 
            \caption{with ToF senbsors} 
            \label{fig:MRtof} 
            \vspace{4ex}
        \end{subfigure}
        \caption{Two variants of MiniRyś mobile robot}
        \label{fig:MRversions} 
    \end{figure}

    The first prototypes of the MiniRyś mobile robot were developed as small-sized platforms [R7] based on research on a~large-scale platform with variable locomotion mode \cite{kkr15_minirys-eng}. In subsequent developmental versions of the robot, individual hardware components, software solutions, and operating systems were investigated. Currently, the robot features a~control system based on ROS 2 framework and MeROS metamodel\cite{winiarski2023meros}\footnote{\url{https://github.com/twiniars/MeROS}} [R9], created as part of work on a~real-time hardware-software platform. In the latest work on the robot's newest version, emphasis was placed on diagnostics, tests, and operation visualization \cite{jostrysz-bsc-23-eng}. The robot's hardware platform is equipped with a~rotating LiDAR scanner (fig. \ref{fig:MRlidar}), and the control system has been integrated with the Nav2 navigation stack. In addition to moving in horizontal mode, where the bumper serves as the third point of support, it also can move in vertical mode, where the robot balances on two wheels. The robot also has versions equipped with ToF sensors, allowing it to move and navigate vertically in narrow corridors (fig. \ref{fig:MRtof}).

     The robot serves as an educational and research platform, enabling training in the control of mobile robots, vision systems, and real-time systems, as well as conducting research on managing multi-robot systems and swarm robot behaviour.

	\section{Selected applications}
	\label{sec:applications}
	
	\subsection{TwitchGo}
    \label{subsec:twitchgo}
	
    TwitchGo project is a~student-driven initiative that aims to research robotic systems controlled simultaneously by local and remote controllers by emulating a~rivalry-driven game. During the game, two participants standing directly next to the board drive separate Homar robots (sec. \ref{subsec:homar_mobile_robot}) and compete by gathering small balls from the board to their respective goals. The event is transmitted live on the Twitch platform as a~match. Simultaneously, the viewers can send specific comments in the stream's chat to activate mechanised obstacles placed in the board (sec. \ref{subsec:mechanised_obstacles}). By doing this, they choose to help both, either or neither of the teams.
	
	\subsection{Line and corridor follower}
	    
\begin{figure}[H]
        \centering
        \begin{subfigure}[b]{0.48\textwidth}
            \includegraphics[width=\textwidth]{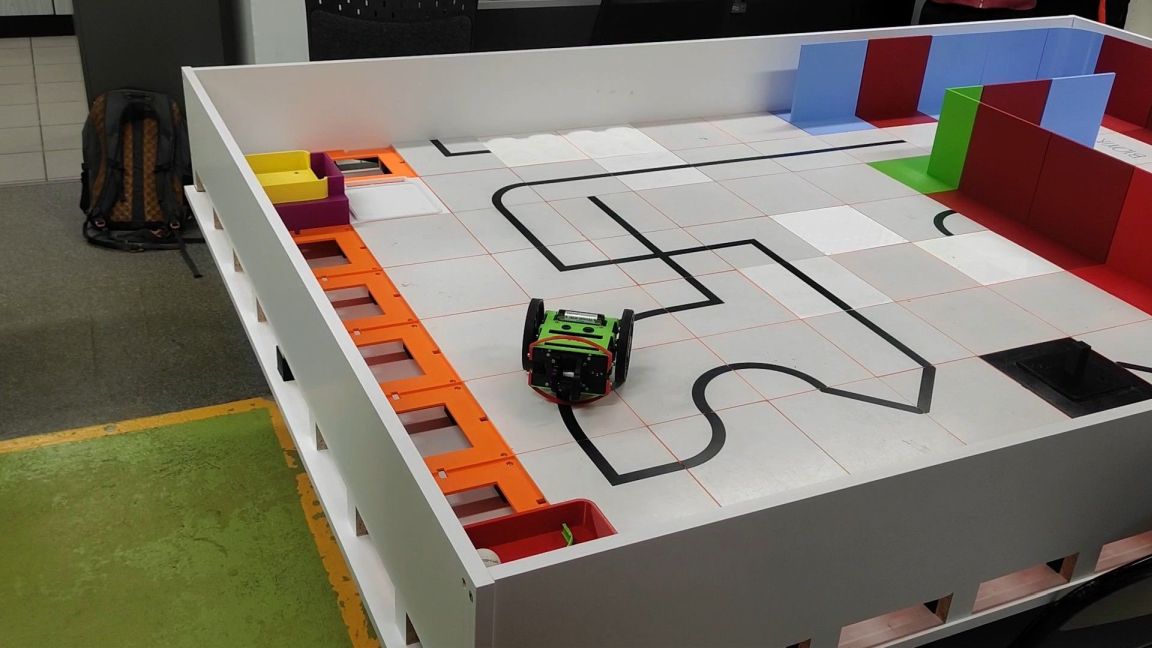}
            \caption{Line following}
            \label{fig:linefollo}
        \end{subfigure}
        \hfill
        \begin{subfigure}[b]{0.48\textwidth}
            \includegraphics[width=\textwidth]{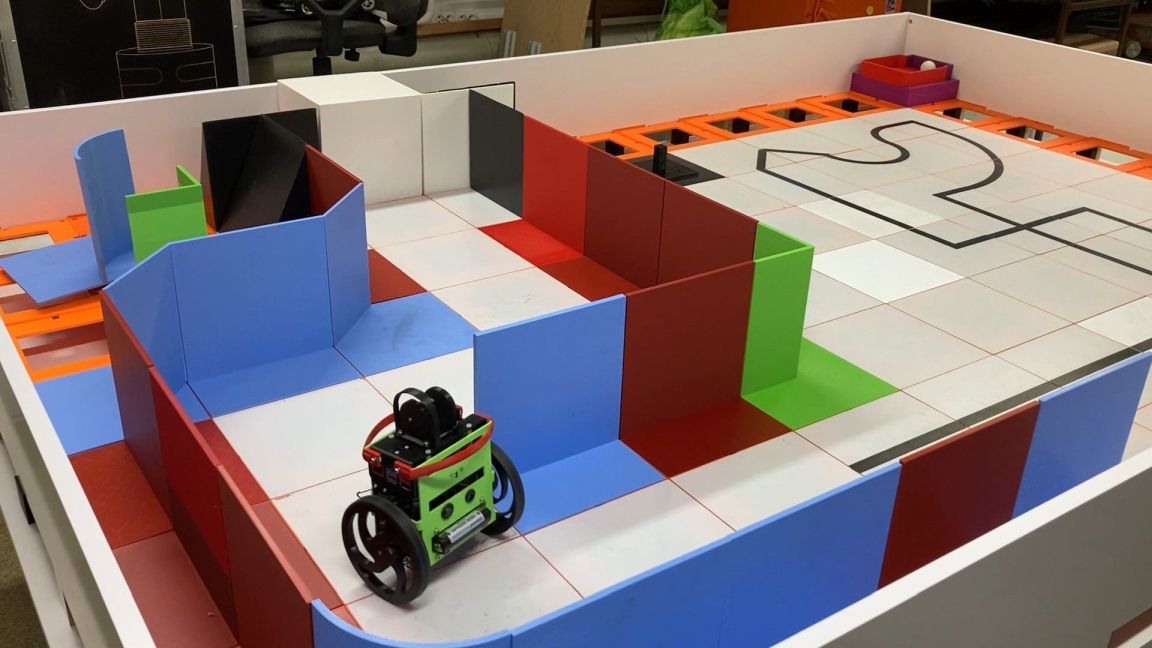}
            \caption{Corridor following}
            \label{fig:wallfollo}
        \end{subfigure}
        \label{fig:minirtraska}
    \end{figure}

	As a~part of Bachelor's thesis~\cite{jsadowski-bsc-24-eng}, the line following horizontally (fig. \ref{fig:linefollo}) and moving through a~corridor (fig. \ref{fig:wallfollo}) vertically by MiniRyś robot has been performed. The first task, related to line following, is based on image analysis from a~built-in camera. The vertical movement task uses the robot's ability to balance, which was achieved by cascade control. Both functions were included in a~single scenario\footnote{\url{https://vimeo.com/901892055}}.
 
	\subsection{Multirobot transport of objects}
	 
\begin{figure}[H]
	\centering
	\includegraphics[height=5cm]{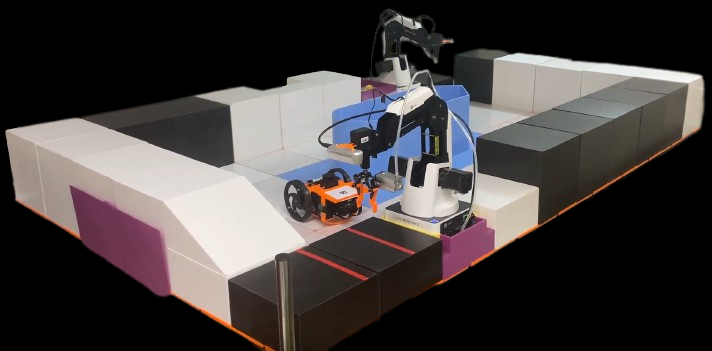}
	\caption{Test environment for robot cooperation in a~transportation task}
	\label{fig:test_env}
\end{figure}

  As a~part of the project \textit{"Development of methods of cooperation between manipulation and mobile robots using the ROS 2 framework"} a~practical validation of the possibility of cooperation of heterogeneous robots in a~transport task and in a~\textit{Pick}\&\textit{Place} task was carried out\footnote{\url{https://vimeo.com/865928183}}. Appropriate modifications were made at the mechanical and software levels of the mobile and manipulation robots. These modifications expanded their abilities to operate in a~dynamically changing environment. A~prototype of a~modular board was designed and manufactured that provided a~suitable operational environment for all participating robots (fig.~\ref{fig:test_env}). The structure and behaviours of the system were modelled in SysML using MeROS metamodel\cite{winiarski2023meros}\footnote{\url{https://github.com/twiniars/MeROS}}.

	\section{Conclusions}
	\label{sec:conclusions}
	
	The HeROS platform has been in development for many years and has been the basis for a~variety of works. In the near future, we plan to focus on exploring different strategies for the execution of multi-robot tasks involving mobile robots and manipulators. Strategies include both coordinated actions and those based on swarm intelligence.

	
	\section*{Acknowledgment}
		
    The authors would like to thank Maciej Bogusz, Mateusz Dumin and Hubert Kowalski for their support in developing the current version of the MiniRyś robot, Paweł Rawicki for the development of the MiniRyś graphical robot control interface, Bartek Kwieciński, Małgorzata Dulikowska and other members of TwitchGO team for the design of the Homar mobile robot and mechanized obstacles. 

    The project was funded by the Rector's grants for Bionik -- Students Robotics Interest Club in 2016-2020, Excellence Initiative: Research University (IDUB) Programme in 2022/2023 and Institute of Control and Computation Engineering.
	
	
	\bibliography{HeROS-arxiv-24}
	\bibliographystyle{pprai}
	
\end{document}